\documentclass[10pt,twocolumn,letterpaper]{article}

\usepackage{wacv}              

\definecolor{wacvblue}{rgb}{0.21,0.49,0.74}
\usepackage[pagebackref,breaklinks,colorlinks,allcolors=wacvblue]{hyperref}

\usepackage{graphicx}
\usepackage{placeins}
\usepackage[english]{babel}
\usepackage{url}
\usepackage{algorithm}
\usepackage{algpseudocode}
\usepackage{xcolor}
\usepackage{array}
\usepackage{bm} 
\usepackage{multirow}
\usepackage{booktabs} 
\usepackage{amssymb}
\usepackage{amsmath}
\usepackage{caption}
\usepackage{changepage}
\usepackage{booktabs}
\usepackage{colortbl}
\usepackage{color, colortbl}
\usepackage{arydshln}
\usepackage{booktabs}
\usepackage{hyperref}
\usepackage{wrapfig}

\definecolor{mygray}{rgb}{.95,.95,.95}

\definecolor{Gray}{gray}{0.9}

\newcolumntype{x}[1]{>{\arraybackslash\hspace{0pt}}p{#1}}

\def\wacvPaperID{1417} 
\def\confName{WACV}
\def\confYear{2026}

\title{EFSA: Episodic Few-Shot Adaptation for Text-to-Image Retrieval}

\author{Muhammad Huzaifa\\
Institution1\\
Institution1 address\\
{\tt\small firstauthor@i1.org}
\and
Yova Kementchedjhieva \\
Institution1\\
Institution1 address\\
}

\author{Muhammad Huzaifa$^{\dagger}$ \ \ Yova Kementchedjhieva$^{\dagger}$ \\
        $^{\dagger}$Department of Natural Language Processing, MBZUAI \\ 
        \texttt{\{muhammad.huzaifa, yova.kementchedjhieva\}@mbzuai.ac.ae}}

\begin{document}
\maketitle
\begin{abstract}

Text-to-image retrieval is a critical task for managing diverse visual content, but common benchmarks for the task rely on small, single-domain datasets that fail to capture real-world complexity. Pre-trained vision-language models tend to perform well with easy negatives but struggle with hard negatives—visually similar yet incorrect images—especially in open-domain scenarios. To address this, we introduce Episodic Few-Shot Adaptation (EFSA), a novel test-time framework that adapts pre-trained models dynamically to a query’s domain by fine-tuning on top-$k$ retrieved candidates and synthetic captions generated for them. EFSA improves performance across diverse domains while preserving generalization, as shown in evaluations on queries from eight highly-distinct visual domains and an open-domain retrieval pool of over one million images. Our work highlights the potential of episodic few-shot adaptation to enhance robustness in the critical and understudied task of open-domain text-to-image retrieval.

\end{abstract}    
\section{Introduction}
\label{sec:intro}

One of the key features of foundation vision-language models, such as CLIP \cite{radford2021learningtransferablevisualmodels}, is their ability to perform text-to-image retrieval: a task that enables humans and systems to efficiently sift through the vast and ever-growing amounts of visual information produced daily through photography, art, graphic design, scientific imaging, and more. As fundamental and all-encompassing as this task is, the most commonly-used evaluation benchmarks for it, COCO \cite{lin2015microsoft} and Flickr30k \cite{plummer2015flickr30k}, hardly do it justice. These datasets rely on small pools of candidate images (5k and 1k, respectively), both of which concern a single visual domain: natural photos, and exhibit limitations in representing even that domain sufficiently well \cite{zhao2017men, burns2018women}. Evaluation protocols that only rely on this data—which is not uncommon in vision-language research \cite{lu2019vilbert, li2020oscar, yao2021filip}—likely overestimate the capabilities of vision-language models and obfuscate the gap in performance on open-domain text-to-image retrieval. 

\begin{figure}[!t]
    \centering
    \includegraphics[width=\linewidth]{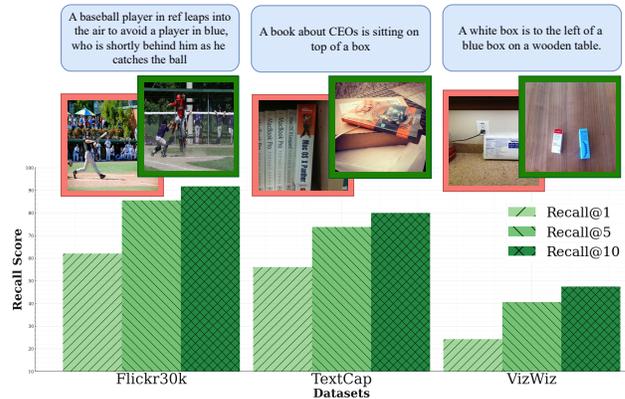}
    \caption{Zero-shot text-to-image retrieval with CLIP exhibits a sharp drop in Recall@1 compared to Recall@5 and 10. For the three example queries on top, CLIP ranks an incorrect image (red frame) as the highest, which is highly similar to the ground truth image (green frame). Recall@1 suffers due to such hard negatives.}
    \label{fig:intro}
\end{figure}

A key observation about models like CLIP and SigLIP \cite{zhai2023sigmoid} is that they excel at distinguishing and identifying easy negatives—dissimilar images for text-to-image retrieval tasks, as reflected in their strong performance for higher recall metrics such as Recall@5 and Recall@10 (see Figure \ref{fig:intro} for results on three distinct datasets). However, they often struggle to correctly rank hard negatives, which are images with subtle similarities or minor variations compared to the query, affecting Recall@1. This limitation becomes especially problematic in an open-domain setting where visually similar images across domains may vary slightly, thus requiring more fine-grained alignment.

    


In this work, we consider a more realistic evaluation setting in which images are retrieved from an extensive and diverse pool of over one million candidates, encompassing eight known and highly distinct visual domains as well as several additional unknown domains sourced from a general web-scraped image repository \cite{changpinyo2021conceptual}. We observe that each of the newly added domains poses a greater challenge to CLIP than the natural photos in COCO and Flickr30k (in a single-domain setting), and that retrieval from an open-domain pool is significantly more challenging than retrieval from a single-domain pool. This underscores the need to improve CLIP’s performance across various domains beyond natural images. However, in an open-domain setting, this cannot be achieved through finetuning alone, as any domain left out during finetuning, but present at inference time, would likely suffer from reduced generalization \cite{wortsman2022robust}.

To address these limitations, we introduce a novel framework for open-domain text-to-image retrieval called \textbf{Episodic Few-Shot Adaptation (EFSA)}. EFSA enables a pre-trained vision-language model to adapt dynamically to the specific domain or micro-domain that a query demands at inference. This adaptation process involves retrieving the top-$k$ images most relevant to the query, fine-tuning the model using these images and their synthetically generated captions, and then re-ranking the images with the updated model. This episodic adaptation resets the model parameters after each test sample, ensuring preservation of the generalization acquired during pre-training, for optimal adaptation to each new query. Our evaluations and ablations highlight EFSA’s strengths in enhancing retrieval performance and improving adaptability to a variety of visual domains.

\noindent Our contributions can be summarized as follows:
\begin{itemize}
    \item We identify limitations in current text-to-image retrieval benchmarks, emphasizing the need for evaluation in a realistic, open-domain setting of diverse visual content.
    \item We introduce EFSA, a novel test-time adaptation framework which enhances model robustness against hard negatives, by episodically learning from them.  
    \item We show quantitative and qualitative results demonstrating the robustness of ESFA against hard negatives across a range of eight highly distinct domains. 
\end{itemize}

\noindent The findings of this work suggest that adaptation methods are particularly effective for addressing hard negatives. This substantiates the potential of few-shot adaptation techniques to enhance robustness and generalization in open-domain text-to-image retrieval. 

\section{Background \& Related Work}
\label{sec:related_work}
\subsection{Text-to-Image Retrieval}

\textbf{Definition} Text-to-image (T2I) retrieval involves retrieving relevant images from a predefined pool based on a text query. Formally, let $\mathcal{D} = \{(X_i, T_i)\}_{i=1}^{N}$ represent a dataset of image-text pairs, where $X_i$ is an image and $T_i$ is its corresponding text. Given a text query $T_q$, the goal is to identify the most relevant image $X$ from $\mathcal{D}$ that aligns with $T_q$ in semantic content. This retrieval process relies on encoding both text and image data into a shared representation space, enabling similarity-based matching.

\vspace{2 mm}
\noindent\textbf{Evaluation Datasets} The two most widely used benchmarks for T2I retrieval are COCO \cite{lin2015microsoft} and Flickr30k \cite{plummer2015flickr30k}. COCO is based on 80 object classes originally derived from ImageNet \cite{deng2009imagenet}, focusing on various everyday objects. Flickr30k is designed to capture people engaged in everyday activities and events, and has been shown to largely overlap with COCO in terms of domain coverage \cite{young2014image, chen2015microsoft}.
Both datasets were originally conceived as image-captioning datasets but have since become standard benchmarks for single-domain retrieval evaluation. 

Another dataset, considerably less widely used in T2I evaluations, is VizWiz \cite{gurari2018vizwiz}, which consists of close-up photos taken by visually impaired users in their environment.  

The sole benchmark designed to reflect the true complexity of multi-domain T2I retrieval is P9D \cite{zhu2023ctp}, a dataset of images from nine e-commerce domains, paired with captions in Chinese. Our inspection of these captions revealed them to be rather noisy and underspecific.     

\vspace{2 mm}
\noindent\textbf{Limitations of COCO and Flickr30k}  Numerous works rely heavily on COCO and Flickr30k for T2I retrieval evaluation, underscoring their central role as benchmarks. Models like ViLBERT \cite{lu2019vilbert}, OSCAR \cite{li2020oscar}, and FILIP \cite{yao2021filip} use these datasets extensively to assess natural image retrieval capabilities. Approaches such as ALIGN \cite{jia2021scaling}, Florence \cite{yuan2021florence}, VSE++ \cite{faghri2017vse++}, SCAN \cite{lee2018stacked}, and VisualBERT \cite{li2019visualbert} also prioritize COCO and Flickr30k,\footnote{Just X of these works evaluate retrieval performance on VizWiz.} establishing these datasets as de facto standards for T2I retrieval, despite their limited diversity and narrow domain focus. 

Although these datasets are valuable for evaluating model capabilities in the natural image domain, their limitations in terms of size and coverage hinder a comprehensive evaluation of model performance. Our work addresses this gap by expanding the scope of retrieval evaluation to multiple domains, to assess model performance in a more realistic and challenging open-domain scenario.

\begin{figure*}[t]
    \centering
    \includegraphics[width=\textwidth]{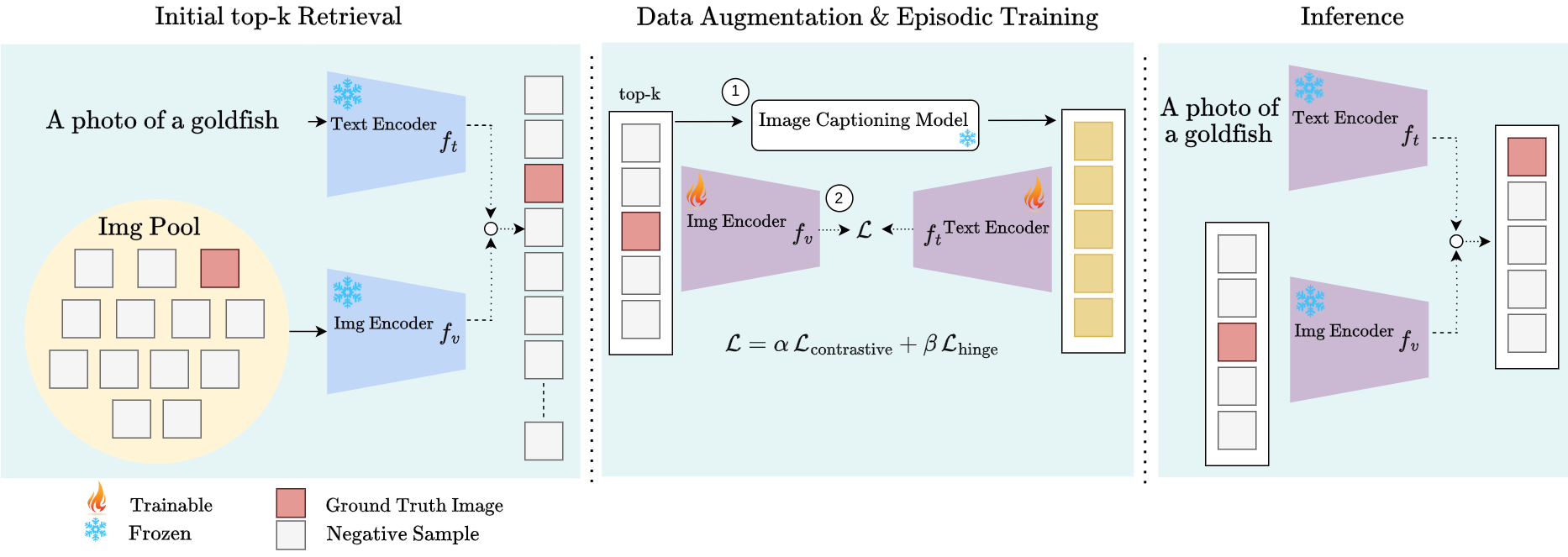}
    \caption{Our method, Episodic Few-Shot Adaptation, works by first retrieving the top-$k$ most similar images from a diverse, open-domain image pool. It then finetunes both the image and text encoder on these top-$k$ images and synthetic captions generated for them. Finally, the updated encoders are used to re-rank the top-$k$ images, bringing more correct candidates to the high ranks.}
    \label{fig:approach}
\end{figure*}

\subsection{Adaptation Methods}

Episodic training, a key component of few-shot learning \cite{vinyals2016matching, snell2017prototypical}, has proven effective for handling diverse and unseen conditions by enabling task specification without loss of generalization \cite{finn2017model, ravi2017optimization}. This approach improves model adaptability by structuring training into discrete episodes, each designed as a small-scale task or domain-specific learning session. Building on these principles, test-time adaptation (TTA) enables models to adjust dynamically to distribution shifts encountered during inference, fine-tuning their parameters on each test sample \cite{sun2020test, wang2020tent, ttt_new}. TTA has proven effective in enhancing the robustness of vision-language models, helping them handle both in-domain and out-of-domain variations. Numerous recent studies employ TTA   successfully for the task of image classification but not to retrieval \cite{shu2022test, guo2023calip, feng2023diverse, hassan2024align, imam2024test}. 
Recently, RLCF \cite{zhao2024testtime} used a teacher-student framework to guide adaptation for classification as well as retrieval, evaluating yet again on just COCO and Flickr30k, in a single-domain setting. RLCF serves as a key baseline in our study. 

Continual learning also addresses multi-domain scenarios, by incrementally tuning a model to new domains to mitigate catastrophic forgetting \cite{zhu2023ctp}. Yet, this approach assumes knowledge of the relevant domains and access to training data, both impractical for real-world T2I retrieval.

\subsection{Synthetic Captions for Retrieval}

Recent work has shown that continual pre-training of VLMs like CLIP on synthetically generated captions yields improved performance on a range of tasks, including retrieval \cite{Chen2023ShareGPT4VIL, Zhang2024LongCLIPUT, Wu2024LoTLIPIL}. Using a generative VLM like LLaVa \cite{liu2023llava}, these works re-label millions of images with well-formed detailed captions, far more informative than the noisy captions used for the initial pre-training of CLIP  \cite{radford2021learningtransferablevisualmodels}. While continued pre-training allows the modified VLM to perform better on average across many tasks and domains, it does not guarantee optimal performance in any one of these tasks and domains. Our episodic few-shot training framework leverages just a few highly relevant synthetically captioned images to tune the VLM to the specific domain of the query. 

Iijima \textit{et al.} \cite{iijima2024multimodal} propose an alternative use of synthetic captions as domain-agnostic representations of images and use text-to-text retrieval as a proxy for cross-domain image-to-image retrieval. We include a text-to-text retrieval baseline, finding it to be far weaker than our proposed approach. 

\section{Method}
\label{sec:method}
\subsection{Preliminaries}

\noindent\textbf{Contrastive Language-Image Pre-training (CLIP)} comprises two encoders: the visual encoder $\mathcal{F}_{\theta_v}$, which maps visual input $X$ to a fixed-length representation $\bm{f}_v$, and the text encoder $\mathcal{F}_{\theta_t}$, which processes text input, $T$, and generates a latent textual feature $\bm{f}_t$. 
In zero-shot T2I retrieval, a string of text, $q$,  is used to query a pool of images, $\mathcal{I}$. Given the representations of the query, $\bm{f}_q$, and of all images $\{\bm{f}_{v,i}\}_{i \in \mathcal{I}}$, a cosine similarity score $s_i=\mathtt{sim}(\bm{f}_q, \bm{f}_{v,i})$ is computed for each image, $i$, to obtain a final ranking.



\vspace{2 mm}
\noindent \textbf{Episodic Training} is a learning framework which structures the training process into a sequence of episodes, where each episode is crafted to simulate a distinct task or domain. An episode comprises a support set $\mathcal{S} = \{(X_i, y_i)\}_{i=1}^{N}$, containing labeled instances for learning, and a query set $\mathcal{Q}$, used for evaluation. The model is trained to optimize its performance on $\mathcal{Q}$ based on information from $\mathcal{S}$, encouraging task-specific adaptations within each episode. 


\subsection{EFSA: Episodic Few-Shot Adaptation}

\paragraph{Overview} Although CLIP-style models achieve impressive zero-shot performance across varied downstream tasks, they struggle to accurately rank hard negative pairs in T2I retrieval, particularly in complex, multi-domain scenarios. We address this limitation with EFSA, an adaptation framework which utilizes top-$k$ highly similar images as hard negatives, to improve multi-modal alignment in the specific query domain.
In this section, we detail our three-stage EFSA methodology, explaining how each stage contributes to more robust retrieval across diverse and challenging domains. For a visualization of the approach, see Figure~\ref{fig:approach}.

\vspace{2 mm}
\noindent \textbf{Initial Top-$k$ Retrieval}
In the initial zero-shot retrieval step, we compute the score, $s$, between a given query text, ${q}$, and all images $\bm{i} \in \mathcal{I}$. 
The top-$k$ most similar images are selected, forming the set $\mathcal{I}_{\text{top}} = \{\bm{i}_1, \bm{i}_2, \dots, \bm{i}_k\}$. Given a large-enough $K$ and a sufficiently generalized VLM, we can expect the ground truth to be in $\mathcal{I}_{\text{top}}$, but possibly not at the top rank, where it should be. Crucially, the rest of the images in the set would be similar to each other and to the ground truth, thus forming a set of domain-specific hard negatives with respect to the ground truth. 

\vspace{2 mm}
\noindent \textbf{Data Augmentation and Episodic Training}
In this step, pseudo-captions  $\mathcal{C}_{\text{top}} = \{\bm{c}_{1}, \bm{c}_{2}, \dots, \bm{c}_{k}\}$ are generated for each image in $\mathcal{I}_{\text{top}}$ using a pre-trained image captioning model, guided by a predefined prompt $\mathcal{P}$, chosen to ensure relevant caption generation. 
Together $\mathcal{I}_{\text{top}}$ and $\mathcal{C}_{\text{top}}$ form a dataset of image-caption pairs, which are used to adapt the VLM using two losses: contrastive and hinge.

For a batch of $N$ image-text pairs $(X_i, T_i)$, the contrastive loss is formulated as:

\begin{equation}
\mathcal{L}_{\text{contrastive}} = -\frac{1}{N} \sum_{i=1}^{N} \log \frac{\exp(\text{sim}(\bm{f}_{v,i}, \bm{f}_{t,i}) / \tau)}{\sum_{j=1}^{N} \exp(\text{sim}(\bm{f}_{v,i}, \bm{f}_{t,j}) / \tau)},
\end{equation}

\noindent where $\text{sim}(\cdot, \cdot)$ denotes cosine similarity
, and $\tau$ is the temperature parameter.

The hinge loss further penalizes misalignment by enforcing a margin $m$ between positive and negative pairs, ensuring the distinctness of representations. It is defined as:

\begin{align}
\mathcal{L}_{\text{hinge}} = \frac{1}{N} \sum_{i=1}^{N} \sum_{j \neq i} &\max \left( 0, m - \text{sim}(\bm{f}_{v,i}, \bm{f}_{t,i}) \right. \nonumber \\
&\left. + \text{sim}(\bm{f}_{v,i}, \bm{f}_{t,j}) \right).
\end{align}

The final test-time adaptation loss, $\mathcal{L}_{\text{test}}$, is a weighted combination of the two: $
\mathcal{L}_{\text{test}} = \alpha \, \mathcal{L}_{\text{contrastive}} + \beta \, \mathcal{L}_{\text{hinge}}
$
, where $\alpha$ and $\beta$ balance the contribution of each component.

These objectives are used to learn a set of LoRA layers \cite{hu2021lora} in both $\mathcal{F}_{\theta_v}$ and $\mathcal{F}_{\theta_t}$ in the VLM. With LoRA, the VLMs is adapted at a low computational cost, with reduced risk of overfitting to noise in $\mathcal{I}_{\text{top}}$ and $\mathcal{C}_{\text{top}}$.


\vspace{2 mm}
\noindent \textbf{Inference} In this final step, the updated VLM is used to re-rank the images in $\mathcal{I}_{\text{top}}$ with respect to the query $q$, using cosine similarity and leveraging domain-adapted representations to enhance retrieval recall at the top ranks. 

After each inference step, all LoRA weights are reset, ensuring that subsequent queries start from a neutral state, allowing the model to adapt to each new query independently, without cross-domain interference.
\section{Experiments}
\label{sec:experiments}

\begin{table}[t]
\centering
\caption{Image captioning datasets used for evaluation. We report the source of the data, the split that we use, whether we subsample this split ($\subseteq$), the number of data points used (\#), and the domain.  }
\footnotesize
\resizebox{\columnwidth}{!}{%
\begin{tabular}{lllll}
\hline
\rowcolor{gray!10}\textbf{Dataset} & \textbf{Split} & \textbf{$\subseteq$ } & \textbf{\#} & \textbf{Domain}  \\
\hline
COCO \cite{lin2015microsoft} & Test & $\times$ & 5K & Natural Scenes \\
\rowcolor{gray!10}Flickr30k \cite{plummer2015flickr30k}  & Test & $\times$ & 1K & Natural Scenes  \\
Books \cite{oldbookillustrations_2007}  & Train & $\times$ & 4.1K & Illustrations  \\
\rowcolor{gray!10}NASA Earth \cite{nasa_earth_instagram_hf}  & Train & $\times$ & 415 & Satellite Imagery  \\
ArtCap \cite{9965360} & Test & $\times$ & 3.6K & Fine Art Paintings  \\
\rowcolor{gray!10}SciCap \cite{hsu2021scicapgeneratingcaptionsscientific} & Test & \checkmark & 3K & Scientific Figures  \\
VizWiz \cite{gurari2018vizwiz} & Val. & $\times$ & 7.7K & Assistive Images  \\
\rowcolor{gray!10}TextCaps \cite{sidorov2020textcaps} & Val.  & $\times$ & 3.1K & Images with Text \\
CC12M \cite{changpinyo2021conceptual} & Train & \checkmark & 1M & Open-Domain  \\
\hline
\end{tabular}%
}
\label{tab:dataset}
\end{table}

\begin{table*}[t]
\fontsize{10pt}{5pt}\selectfont
\centering
\caption{
Text-to-image retrieval performance in a single-domain setting. Results are reported for Zero-Shot (Z.S), Fine-Tuning (F.T), Text-to-Text (T2T), RLCF, and Episodic Few-Shot Adaptation (EFSA). EFSA performs best on average on Recall@1.}
\renewcommand{\arraystretch}{1.4}
\resizebox{\linewidth}{!}{%
\begin{tabular}{lccccccccccccccccccccccccccc}
\toprule

& \multicolumn{12}{c}{\textbf{Single-domain}}\\

\midrule 
\multirow{2}{*}{} & \multicolumn{3}{c}{\textbf{COCO}} & \multicolumn{3}{c}{\textbf{Flickr30k}} & \multicolumn{3}{c}{\textbf{Books}} & \multicolumn{3}{c}{\textbf{NASA}} \\ 
\cmidrule(lr){2-4}
\cmidrule(lr){5-7}
\cmidrule(lr){8-10}
\cmidrule(lr){11-13}

& \cellcolor{gray!10}R@1 & R@5 & R@10 & \cellcolor{gray!10}R@1 & R@5 & R@10 & \cellcolor{gray!10}R@1 & R@5 & R@10 & \cellcolor{gray!10}R@1 & R@5 & R@10  \\
\cmidrule(lr){1-13}
 Z.S  & \cellcolor{gray!10}33.07 & 58.42 & 69.00 & \cellcolor{gray!10}62.08 & 85.57 & 91.76 & \cellcolor{gray!10}18.38 & 32.37 & 38.31 & \cellcolor{gray!10}32.53 & 63.13 & 74.21 \\
F.T  & \cellcolor{gray!10}37.35 & 63.35 & 74.19 & \cellcolor{gray!10}\textbf{72.56} & \textbf{91.64} & \textbf{95.35} & \cellcolor{gray!10}14.16 & 27.81 & 33.77 & \cellcolor{gray!10}32.77 & 63.85 & 73.73 \\
T2T & \cellcolor{gray!10}23.12 & 42.53 & 51.90 & \cellcolor{gray!10}37.65 & 57.16 & 64.16 & \cellcolor{gray!10}2.43 & 6.41 & 10.01 & \cellcolor{gray!10}7.71 & 22.16 & 30.6 \\
RLCF & \cellcolor{gray!10}33.72 & 59.14 & 69.78 & \cellcolor{gray!10}63.04 & 86.54 & 92.50 & \cellcolor{gray!10}19.20 & 33.36 & \textbf{39.56} & \cellcolor{gray!10}4.33  & 9.64  & 17.60 \\
\rowcolor{blue!15} EFSA & \textbf{40.41} & \textbf{65.01} & \textbf{72.89} & 68.98 & 89.48 & 93.68 & \textbf{19.71} & \textbf{33.72} & 38.86 & \textbf{34.94} & \textbf{65.78} & \textbf{75.18} \\

\midrule

\multirow{2}{*}{} & \multicolumn{3}{c}{\textbf{VizWiz}} & \multicolumn{3}{c}{\textbf{TextCap}} & \multicolumn{3}{c}{\textbf{ArtCap}} & \multicolumn{3}{c}{\textbf{SciCap}} & \multicolumn{3}{c}{\textbf{Average}}   \\ 
\cmidrule(lr){2-4}
\cmidrule(lr){5-7}
\cmidrule(lr){8-10}
\cmidrule(lr){11-13}
\cmidrule(lr){14-16}

& \cellcolor{gray!10}R@1 & R@5 & R@10 & \cellcolor{gray!10}R@1 & R@5 & R@10 & \cellcolor{gray!10}R@1 & R@5 & R@10 & \cellcolor{gray!10}R@1 & R@5 & R@10 & \cellcolor{gray!10}R@1 & R@5 & R@10 \\
\midrule

Z.S & \cellcolor{gray!10}24.27 & 40.63 & 47.47 & \cellcolor{gray!10}56.08 & 73.81 & 80.04 & \cellcolor{gray!10}15.44 & 32.98 & 42.46 & \cellcolor{gray!10}17.86 & 27.50 & 31.90 & \cellcolor{gray!10}32.46 & 51.80 & 59.39\\
F.T & \cellcolor{gray!10}27.85 & \textbf{45.96} & \textbf{53.35} & \cellcolor{gray!10}\textbf{61.66} & \textbf{79.22} & \textbf{84.41} & \cellcolor{gray!10}\textbf{21.53} & \textbf{43.41} & \textbf{68.58} & \cellcolor{gray!10}19.20 & 29.13 & 34.43 & \cellcolor{gray!10}35.88 & \textbf{55.54} & \textbf{64.72} \\
T2T & \cellcolor{gray!10}16.51 & 28.96 & 34.46 & \cellcolor{gray!10}27.01 &40.99  & 48.00 & \cellcolor{gray!10}6.75 & 14.98 & 19.61 & \cellcolor{gray!10}5.99 & 10.49 & 13.30 & \cellcolor{gray!10}15.89 & 27.96 & 34.00 \\
RLCF & \cellcolor{gray!10}25.00 & 41.41 & 48.29 & \cellcolor{gray!10}11.63 & 57.95 & 65.56 & \cellcolor{gray!10}16.05 & 34.14 & 43.69 & \cellcolor{gray!10}\textbf{24.40} & \textbf{36.80} & \textbf{42.00} & \cellcolor{gray!10}24.67 & 44.87 & 52.37 \\
\rowcolor{blue!15} EFSA & \textbf{28.39} & 44.59 & 49.72 & 61.01 & 76.98 & 81.70 & 19.93 & 38.49 & 45.91 & 20.53 & 30.00 & 33.39 & \textbf{36.73} & 55.50 & 61.41 \\

\bottomrule
\end{tabular}%
}
\label{tab:single-domain}
\end{table*}

\subsection{Experimental Setup}

\noindent\textbf{Implementation Details} We use CLIP-B16 as the main VLM in all experiments, but additionally test the generalization of our approach to SigLIP \cite{zhai2023sigmoid} as well (see Supplementary \S\ref{sec:siglib}). LoRA layers are added to all attention layers and multi-layer perceptron components, in both the vision and text encoders.
The LoRA weights are initialized with Xavier initialization, using a rank of $r = 64$ and a scaling factor of 15. The loss function is updated in a single step with parameters $\alpha = 1.7$ and $\beta = 0.3$, using the AdamW optimizer and a learning rate of 5e-4. All hyperparameters were tuned on the validation splits of COCO and Flickr30k, averaging scores over the two datasets.

For top-$k$ sampling, we set $k=16$. Pseudo-captions are generated with the LLaVA 1.5-13B model \cite{liu2023llava}, using the prompt: \textit{``Describe what you see in detail with a maximum of 30 words."} 
An ablation over different prompts is reported in Supplementary \S\ref{supp:prompts}.
As the captions are independent of the queries, they can be generated offline and cached. Similarly, we precompute and cache the representations of all images in the retrieval pool, as every initial top-$k$ retrieval step relies on the same pre-trained VLM representations. 

All experiments were run for one optimization step (epoch) on a single NVIDIA A100 40GB GPU card.

\vspace{2 mm}
\noindent\textbf{Datasets}  We experiment with a diverse range of image captioning datasets spanning eight domains, detailed in Table~\ref{tab:dataset}. 
We further sample 1 million images sampled from the Conceptual Captions 12M (CC12M) dataset \cite{changpinyo2021conceptual} to represent large-scale open-domain imagery.

\vspace{2 mm}
\noindent\textbf{Experimental Settings}
We study two experimental settings: a \textit{single-domain setting}, where the pool contains only images from a single dataset; and a more challenging \textit{multi-domain setting}, where all datasets in Table~\ref{tab:dataset} are mixed into one large pool. The evaluation sets (e.g., COCO, Flickr30k) remain unchanged, but queries $\mathit{X}$ from domain $\mathit{Y}$ must retrieve their ground-truth among distractors from other domains $\mathit{Y'}$, simulating realistic open-domain conditions.
Images from datasets like CC12M serves solely as distractors, simulating realistic open-domain conditions where the system must retrieve relevant items from a large, diverse, and noisy collection. We recognize that enlarging the retrieval pool may introduce semantically relevant images not labeled as ground-truth (i.e., potential false negatives.) However, since all models in our study are evaluated under the same condition, comparisons remain fair. A robust retriever should be able to pick up even a weak signal, if present, and rank the true match higher than other semantically similar images. 


\vspace{2 mm}
\noindent\textbf{Baselines} 
To show how our adaptation improves over the base model, we report zero-shot (\textbf{Z.S}) T2I retrieval results with CLIP. Since the aligned image-caption pairs from $\mathcal{I}$ and $\mathcal{C}$ form a complete synthetic dataset, we test how standard fine-tuning (\textbf{F.T}) of CLIP on this dataset performs. Here, we use LoRA again, with all the same hyperparameters as defined above, and train the model for four epochs. 
We also include a text-to-text (\textbf{T2T}) retrieval baseline, in which the query text is matched directly against the synthetic captions in $\mathcal{C}_{top}$, using the text encoder of CLIP to obtain representations, and cosine similarity to rank them.  

As an external baseline, we compare against the base variant of RLCF method 
, which uses a CLIP-B16 backbone and a CLIP-L14 model as the teacher. For consistency, we run this method for a single optimization step.

Below, we compare our approach to these baselines using the standard retrieval metric Recall@$k$\footnote{We use Recall@$k$ following prior work~\cite{faghri2017vse++,lee2018stacked,li2019visualbert,zhao2024testtime}. For single ground truth, precision@$k$ is approximately equivalent to Recall@$k$/$k$, and mAP is better suited for multi-target scenarios, which is not the case here.} for $k=1,5,10$, with special attention to Recall@1, since EFSA focuses on an optimal prediction in the top-$1$ position.

\subsection{Results}
\label{sec:results}

\begin{table*}[ht]
\fontsize{10pt}{5pt}\selectfont
\centering
\caption{
Text-to-image retrieval performance in a multi-domain setting. Results are reported for Zero-Shot (Z.S), Fine-Tuning (F.T), Text-to-Text (T2T), RLCF, and Episodic Few-Shot Adaptation (EFSA). EFSA consistently surpasses other methods, particularly on Recall@1.}
\renewcommand{\arraystretch}{1.4}
\resizebox{\linewidth}{!}{%
\begin{tabular}{lccccccccccccccccccccccccccc}
\toprule

& \multicolumn{12}{c}{\textbf{Multi-domain}}\\
\midrule

\multirow{2}{*}{} & \multicolumn{3}{c}{\textbf{COCO}} & \multicolumn{3}{c}{\textbf{Flickr30k}} & \multicolumn{3}{c}{\textbf{Books}} & \multicolumn{3}{c}{\textbf{NASA}} \\ 
\cmidrule(lr){2-4}
\cmidrule(lr){5-7}
\cmidrule(lr){8-10}
\cmidrule(lr){11-13}
& \cellcolor{gray!10}R@1 & R@5 & R@10 & \cellcolor{gray!10}R@1 & R@5 & R@10 & \cellcolor{gray!10}R@1 & R@5 & R@10 & \cellcolor{gray!10}R@1 & R@5 & R@10 \\
\cmidrule(lr){1-13}

Z.S & \cellcolor{gray!10}22.79 & 43.98 & 53.66 & \cellcolor{gray!10}37.05 & 59.34 & 68.50 & \cellcolor{gray!10}10.42 & 20.24 & 25.57 & \cellcolor{gray!10}30.36 & 59.51 & 71.32 \\
F.T & \cellcolor{gray!10}14.98 & 31.82 & 41.26 & \cellcolor{gray!10}21.63 & 41.84 & 51.04 & \cellcolor{gray!10}1.76  & 4.39  & 5.86  & \cellcolor{gray!10}6.02  & 11.80 & 17.83 \\
T2T & \cellcolor{gray!10}18.90 & 35.39 & 43.55 & \cellcolor{gray!10}21.94 & 35.76 & 41.85 & \cellcolor{gray!10}1.36 & 2.87 & 3.98 & \cellcolor{gray!10}4.81 & 12.53 & 17.10 \\
RLCF & \cellcolor{gray!10}22.12 & 42.12 & 51.78 & \cellcolor{gray!10}35.98 & 58.88 & 67.24 & \cellcolor{gray!10}\textbf{10.86}  & \textbf{20.89} & \textbf{26.03} & \cellcolor{gray!10}6.50 & 10.84 & 18.60 \\
\rowcolor{blue!15} EFSA & \textbf{30.14} & \textbf{50.96} & \textbf{57.82} & \textbf{45.71} & \textbf{66.32} & \textbf{71.49} & 10.56 & 19.51 & 24.72 & \textbf{31.08} & \textbf{62.65} & \textbf{72.28} \\

\midrule

\multirow{2}{*}{} & \multicolumn{3}{c}{\textbf{VizWiz}} & \multicolumn{3}{c}{\textbf{TextCap}} & \multicolumn{3}{c}{\textbf{ArtCap}} & \multicolumn{3}{c}{\textbf{SciCap}} & \multicolumn{3}{c}{\textbf{Average}}   \\ 
\cmidrule(lr){2-4}
\cmidrule(lr){5-7}
\cmidrule(lr){8-10}
\cmidrule(lr){11-13}
\cmidrule(lr){14-16}

& \cellcolor{gray!10}R@1 & R@5 & R@10 & \cellcolor{gray!10}R@1 & R@5 & R@10 & \cellcolor{gray!10}R@1 & R@5 & R@10 & \cellcolor{gray!10}R@1 & R@5 & R@10 & \cellcolor{gray!10}R@1 & R@5 & R@10\\
\midrule
Z.S & \cellcolor{gray!10}18.32 & 32.65 & 39.33 & \cellcolor{gray!10}42.94 & 58.29 & 63.97 & \cellcolor{gray!10}7.36  & 17.96 & 24.93 & \cellcolor{gray!10}17.13 & 26.56 & 30.96 & \cellcolor{gray!10}23.29 & 39.81 & 47.28 \\ 
F.T & \cellcolor{gray!10}10.53 & 21.25 & 27.17 & \cellcolor{gray!10}25.77 & 41.52 & 48.62 & \cellcolor{gray!10}4.66  & 12.08 & 17.31 & \cellcolor{gray!10}2.99  & 5.86  & 7.69 & \cellcolor{gray!10}11.04 & 21.32 & 27.09 \\
T2T & \cellcolor{gray!10}14.78 & 26.25 & 31.58 & \cellcolor{gray!10}21.67 & 32.86 & 38.68 & \cellcolor{gray!10}4.05 & 9.14 & 12.03 & \cellcolor{gray!10}5.20 & 8.69 & 10.76 & \cellcolor{gray!10}11.58 & 20.43 & 24.94 \\
RLCF & \cellcolor{gray!10}18.60 & 32.10 & 38.71 & \cellcolor{gray!10}9.00 & 50.24 & 57.85 & \cellcolor{gray!10}7.67  & 18.76 & 25.81 & \cellcolor{gray!10}\textbf{23.63} & \textbf{35.80} & \textbf{40.97} & \cellcolor{gray!10}16.30 & 33.16 & 39.90 \\
\rowcolor{blue!15} EFSA & \textbf{23.46} & \textbf{37.51} & \textbf{41.94} & \textbf{48.71} & \textbf{62.35} & \textbf{66.58} & \textbf{11.13} & \textbf{22.83} & \textbf{27.50} & 19.69 & 28.90 & 32.33 & \textbf{27.56} & \textbf{43.87} & \textbf{49.33}\\

\bottomrule
\end{tabular}%
}
\label{tab:multi_domain}
\end{table*}

The main results are shown in Table \ref{tab:single-domain} for the single-domain setting, and Table \ref{tab:multi_domain} for the multi-domain setting, in which the retrieval pool spans multiple diverse domains. 

\vspace{2 mm}
\noindent\textbf{Single-Domain Setting}
While this setting is not the ultimate target of our work, it lays the foundation for the follow-up discussion of the multi-domain results. 
The first thing to note here is the considerable variation in zero-shot scores across the different datasets. The Books, ArtCap and SciCap datasets, in particular, prove far more challenging than COCO and Flickr30k, underscoring the importance of domain diversity in T2I retrieval evaluation. 

Compared to the zero-shot baseline, fine-tuning generally improves performance. On the one hand, it is not surprising that domain-specific training leads to an improvement in a domain-specific evaluation setting: fine-tuning by design enhances the feature representations of in-distribution data. On the other hand, this result can be interpreted as a diagnostic for the quality of the synthetic image captions used for fine-tuning: poor captions would not have led to a performance improvement. However, we do see a drop in performance for the Books dataset in particular. 

The text-to-text baseline proves to be a weak one, underperforming the zero-shot method by a large margin, with the average Recall@1 score dropping by more than 50\%, from 32.46 with zero-shot retrieval to 15.89 with text-to-text retrieval. This observation is not surprising, considering the inherently lossy nature of text compared to images: language is discrete, imprecise and underspecific. We include this baseline to show that the synthetic captions provide guidance in EFSA not merely through matching the contents of the query text, but rather by allowing the model to learn domain-specific features from the pairing of these captions with the top-$k$ most relevant images.

RLCF delivers improvements over the zero-shot baseline for 6 out of 8 datasets, but these improvements are mostly negligible in comparison to those seen with the fine-tuning baseline. The one exception beign SciCap where RLCF indeed proves best overall, yielding a Recall@1 improvement of 6.54 points. Meanwhile, however, for NASA and TextCap, RLCF exhibits diverging behavior, with scores plummeting for these datasets across all Recall@k metrics. This instability in the reinforcement-based optimizaiton of RLCF leaves simple fine-tuning as the strongest baseline on average for us to compare against.

EFSA proves more effective than fine-tuning on half of the datasets, specifically in terms of Recall@1. EFSA has the capacity to capture more nuanced micro-domain distinctions relevant to the specific test query and its related images. Interestingly, while fine-tuning failed to extract useful knowledge from the synthetic captions for Books, EFSA leverages these captions effectively to achieve an improvement over the zero-shot baseline for this dataset. This suggests that the quality of the synthetic captions is good, but perhaps biases in the data distribution render fine-tuning brittle and EFSA more robust, as it selects which subset of the data to focus on for every test query.  
Among all methods included in this evaluation, EFSA is the only one that never underperforms the zero-shot baseline on Recall@$k$. 

\vspace{2 mm}
\noindent\textbf{Multi-Domain Setting}
In this more challenging and realistic setting, we first note the general decline in zero-shot performance compared to the single-domain scenario, demonstrating the increased difficulty of handling diverse data sources within a unified retrieval pool. For COCO, for example, zero-shot Recall@1 drops from 33.07 to 22.79, likely due to interference from Flickr30k images. Highly distinct datasets like NASA and SciCap, on the other hand, show stable behavior across the two experimental settings. 

We find that in this setting, in contrast to earlier observations, fine-tuning categorically underperforms the zero-shot baseline, with Recall@1 dropping by more than 50\% on average. As the training data now spans a mix of domain, fine-tuning is rendered counterproductive. Meanwhile, for the text-to-text baseline, the trend observed in the single-domain setting holds here as well, with performance being substantially worse than the zero-shot baseline. 
RLCF now outperforms the zero-shot baseline on 4 out of 8 datasets, and notably so only for SciCap. This is the strongest baseline on average in this setting, yet it lags 7 points behind the zero-shot baseline in terms of average Recall@1.

In the multi-domain setting, EFSA proves best on 6 out of 8 datasets, with an average Recall@1 4.27 points over the zero-shot baseline (27.56 v. 23.29). This number coincidentally matches exactly the improvement of EFSA over the zeros-shot baseline in the single-domain setting. In light of the absolute drop in scores in the multi-domain setting compared to the single-domain setting, this stable improvement with EFSA indicates that our method is highly robust to noise in the $\mathcal{I}_{top}$ subset: even if some lower-quality candidates get retrieved, EFSA successfully adapts to the domain-specific patterns and ranks better candidates higher. The same trend is observed when EFSA is applied to a SigLIP backbone (Supplementary Table \ref{tab:siglip_retrieval_comparison}).

Overall, we can conclude from the results presented above that EFSA is indeed a highly performant method for text-to-image retrieval with immense potential, both in a single-domain setting, and even more so in a multi-domain setting, where standard fine-tuning proves inadequate. 

\subsection{Qualitative Analysis}

\begin{figure}[t]
    \centering
    \includegraphics[width=\linewidth]{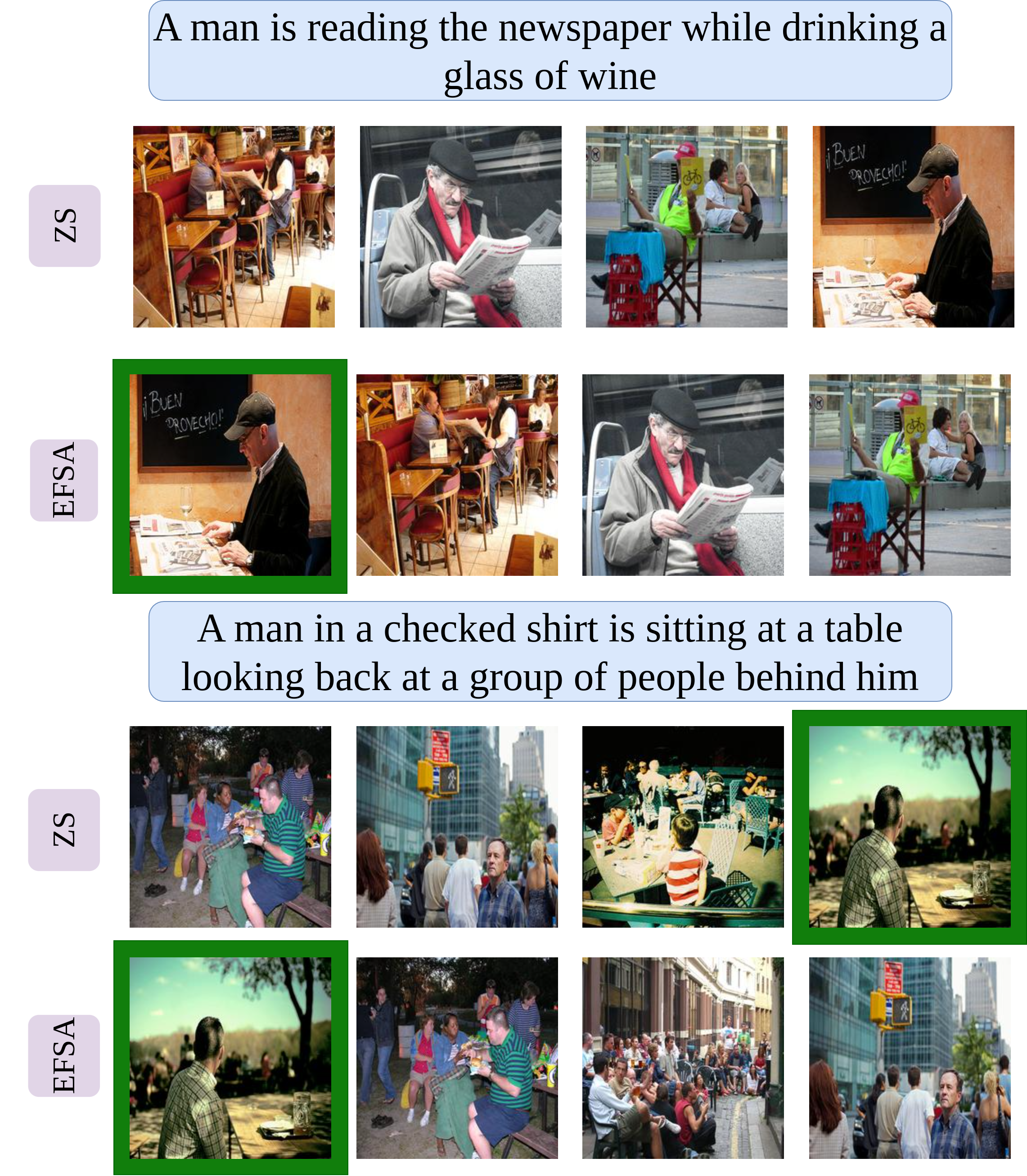}
    \caption{
    Qualitative comparison of EFSA and zero-shot CLIP. Ground-truth images (highlighted in green) correspond to the text queries shown above each row. EFSA successfully re-ranks the ground-truth images to the first rank.
    See Supplementary Figure~\ref{fig:artcap_textcap_sup} for more detailed examples with generated captions.}
    \label{fig:qualitative}
\end{figure}

Figure~\ref{fig:qualitative} provides an insight into the performance gains achieved with EFSA. Looking at the top-4 predictions retrieved with the zero-shot method and with EFSA for two queries from the Flickr30k dataset, we see that EFSA picks up on fine details such as the presence of a wine glass in the top image, and the orientation of the man in the bottom one.

In these examples, we also see evidence for the hard negatives EFSA builds on: in the top example, all pictures show men reading a newspaper, and the image ranked highest by the zero-shot baseline is set in just the right environment for wine consumption. Based on these hard negatives and their synthetic captions, EFSA shifts its focus to the objects and activities present in this few-shot training set.

\subsection{Computational \& Storage Costs}


Like most TTA methods, EFSA introduces additional computation compared to zero-shot retrieval.
Yet, it remains more efficient than prior TTA methods such as RLCF. For example, on the COCO test set (5,000 queries) with $k=8$, EFSA takes 47 minutes, compared to 52 minutes for RLCF in the same setting.  
In the multi-domain setting, as the retrieval pool size increases, the computation scales linearly with the size of the pool, for both the zero-shot retrieval and the initial top-$k$ retrieval step in the TTA methods. In the subsequent adaptation step, however, the computational cost is independent of the pool size and is instead governed by the $k$ parameter.
Since the synthetic captions are generated and cached in advance, there is no added latency from this step at inference time. Although EFSA stores additional image captions compared to the zero-shot baseline, the resulting storage overhead is negligible (see Appendix~\ref{supp:storage}.)

\section{Ablations}


We perform extensive empirical analysis and ablation studies on COCO, NASA, SciCap, and ArtCap to assess the impact of key design choices, including fine-tuning parameters, top-$k$ selection, number of adaptation epochs, and the image captioner. An additional ablation on the loss function is provided in the supplementary material (Appendix~\ref{supp:loss}.)

\subsection{LoRA v. Full Finetuning} 

\begin{figure}[t]
    \centering
    \includegraphics[width=\linewidth]{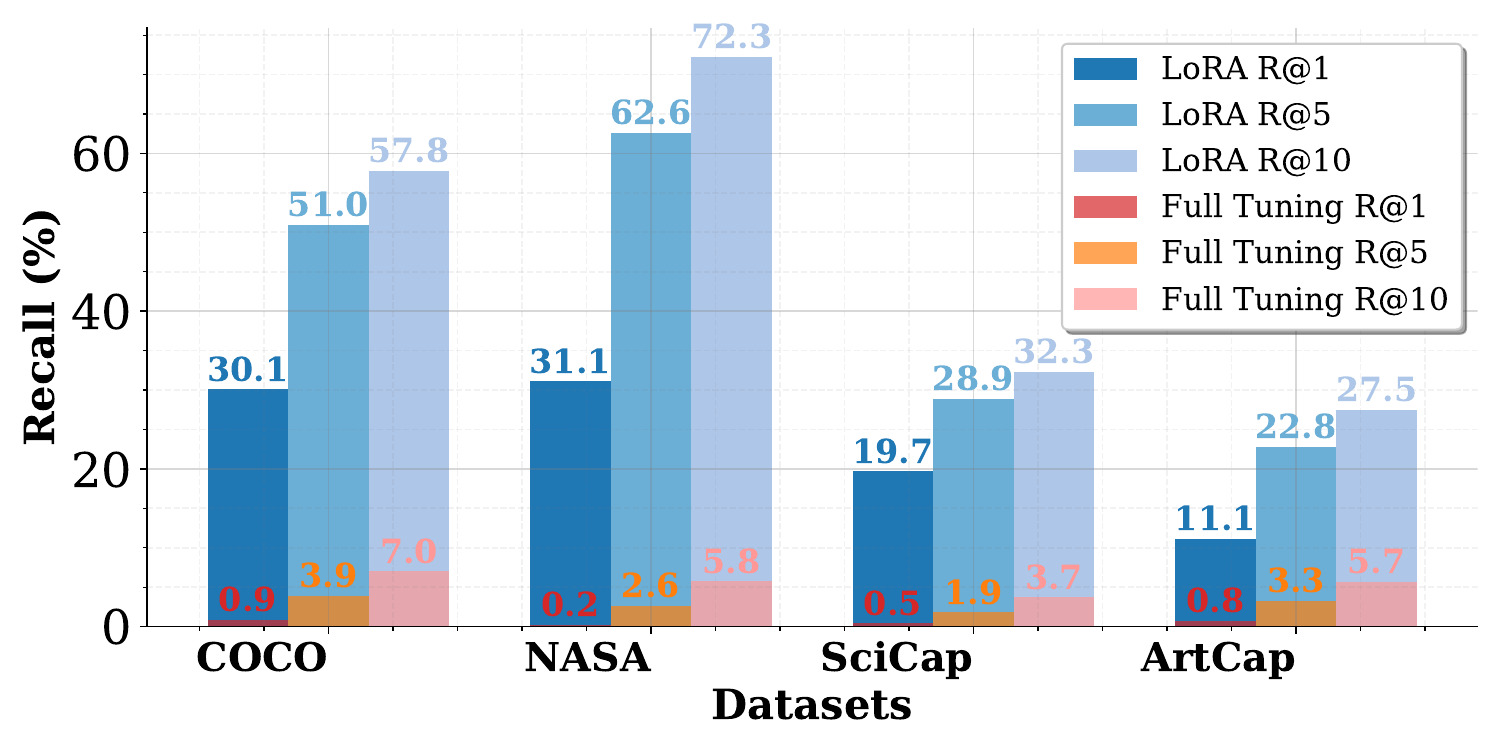}
    \caption{Comparison of LoRA parameter tuning versus tuning all model parameters across multiple datasets}
    \label{fig:lora_full_tuning}
\end{figure}

The results in Figure \ref{fig:lora_full_tuning} highlight the performance differences between LoRA and full parameter tuning across various datasets. LoRA tuning achieves consistently high recall scores, while full parameter tuning falls short across all metrics, demonstrating LoRA's efficiency and suitability for our approach. 
Full tuning typically requires a larger data pool and more training epochs to be effective, as it otherwise lacks the exposure needed to capture a broad range of features. In contrast, LoRA excels in this setup by effectively adapting to one new image-caption pair.


\begin{table*}
\fontsize{5pt}{4pt}\selectfont
\centering
\caption{Comparison of recall performance at incremental top-$k$ values across datasets.}
\renewcommand{\arraystretch}{1.4}
\resizebox{\textwidth}{!}{%
\begin{tabular}{lcccccccccccc}
\toprule
\multirow{2}{*}{\textbf{Top-k}} & \multicolumn{3}{c}{\textbf{COCO}} & \multicolumn{3}{c}{\textbf{NASA}} & \multicolumn{3}{c}{\textbf{SciCap}} & \multicolumn{3}{c}{\textbf{ArtCap}} \\ 
\cmidrule(lr){2-4} \cmidrule(lr){5-7} \cmidrule(lr){8-10} \cmidrule(lr){11-13}
 & \textbf{R@1} & \textbf{R@5} & \textbf{R@10} & \textbf{R@1} & \textbf{@5} & \textbf{R@10} & \textbf{R@1} & \textbf{R@5} & \textbf{R@10} & \textbf{R@1} & \textbf{R@5} & \textbf{R@10} \\ 
\midrule
\rowcolor{gray!10} 8  & 29.83 & 47.50 & 54.44 & 28.91 & 60.24 & 67.95 & 19.63 & 27.56 & 31.00 & 10.75 & 20.75 & 25.36 \\
16 & \textbf{30.14} & 50.96 & 57.82 & 31.08 & 62.65 & 72.28 & \textbf{19.69} & 28.90 & 32.33 & \textbf{11.13} & 22.83 & 27.50 \\
\rowcolor{gray!10} 32 & 30.07 & \textbf{52.11} & 61.05 & \textbf{32.56} & 62.40 &\textbf{ 73.73} & 19.40 & 29.40 & 33.46 & 11.08 & \textbf{24.09} & 30.16 \\
64 & 29.03 & 51.57 & \textbf{61.18} & 32.28 & \textbf{63.85} & 73.01 & 19.56 & \textbf{29.80 }& \textbf{34.20} & 10.64 & 23.86 & \textbf{31.03} \\
\bottomrule
\end{tabular}%
}
\label{tab:topk_recall_comparison}
\end{table*}

\begin{figure*}[t]
    \centering
    \begin{minipage}{0.24\textwidth}
        \centering
        \includegraphics[width=\linewidth]{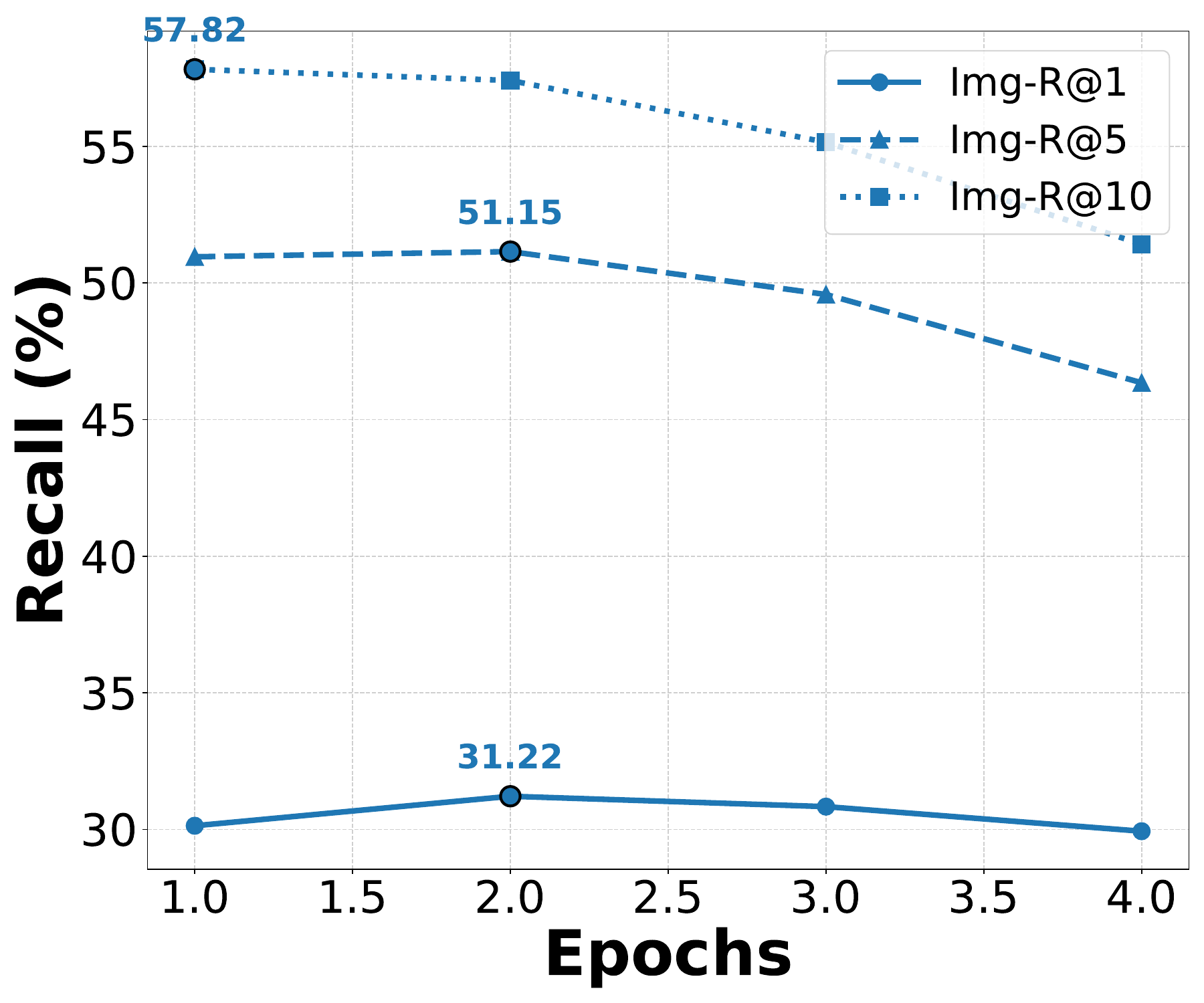}
        \caption*{(a) COCO}
    \end{minipage}
    \begin{minipage}{0.24\textwidth}
        \centering
        \includegraphics[width=\linewidth]{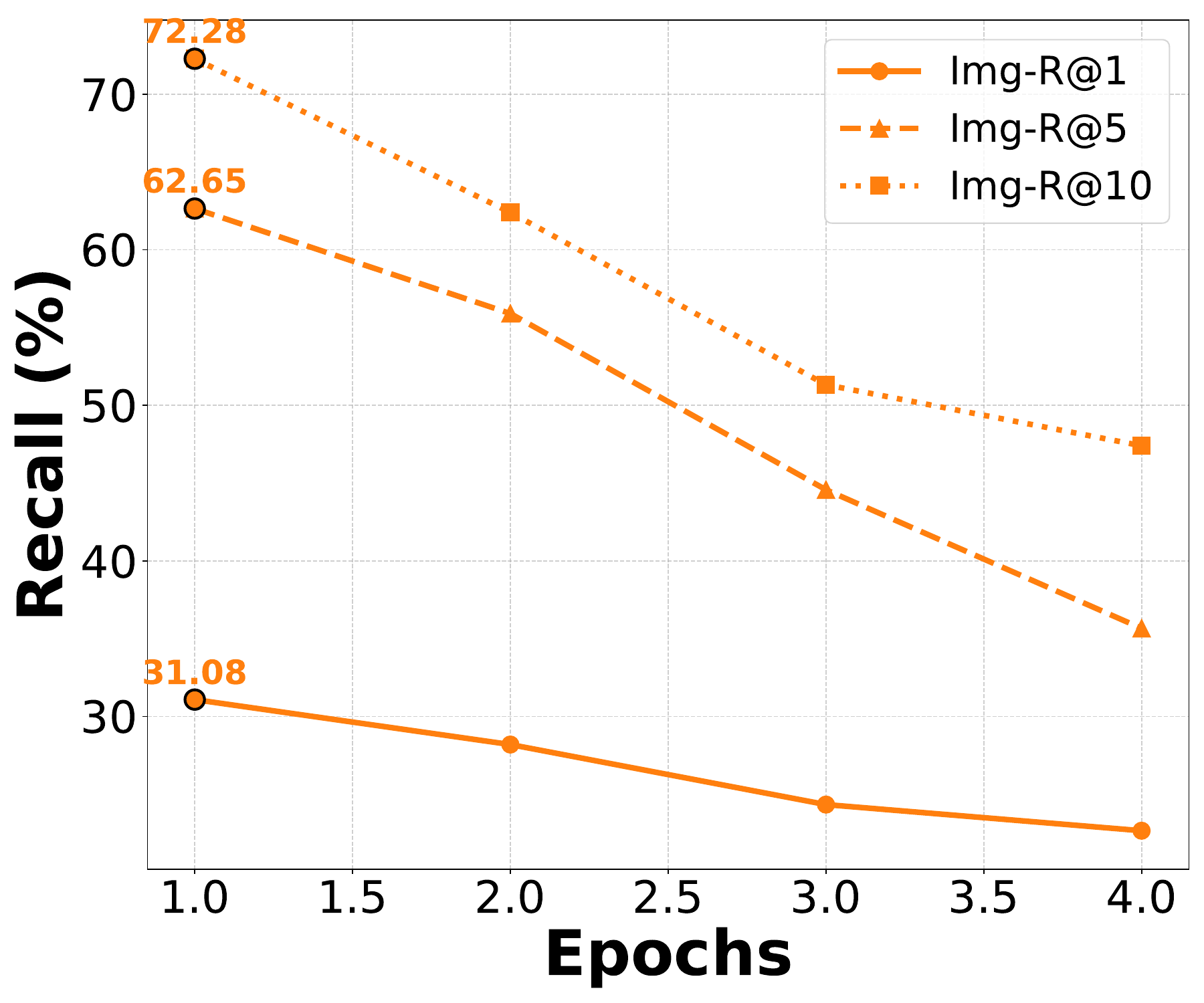}
        \caption*{(b) NASA }
    \end{minipage}
    \begin{minipage}{0.24\textwidth}
        \centering
        \includegraphics[width=\linewidth]{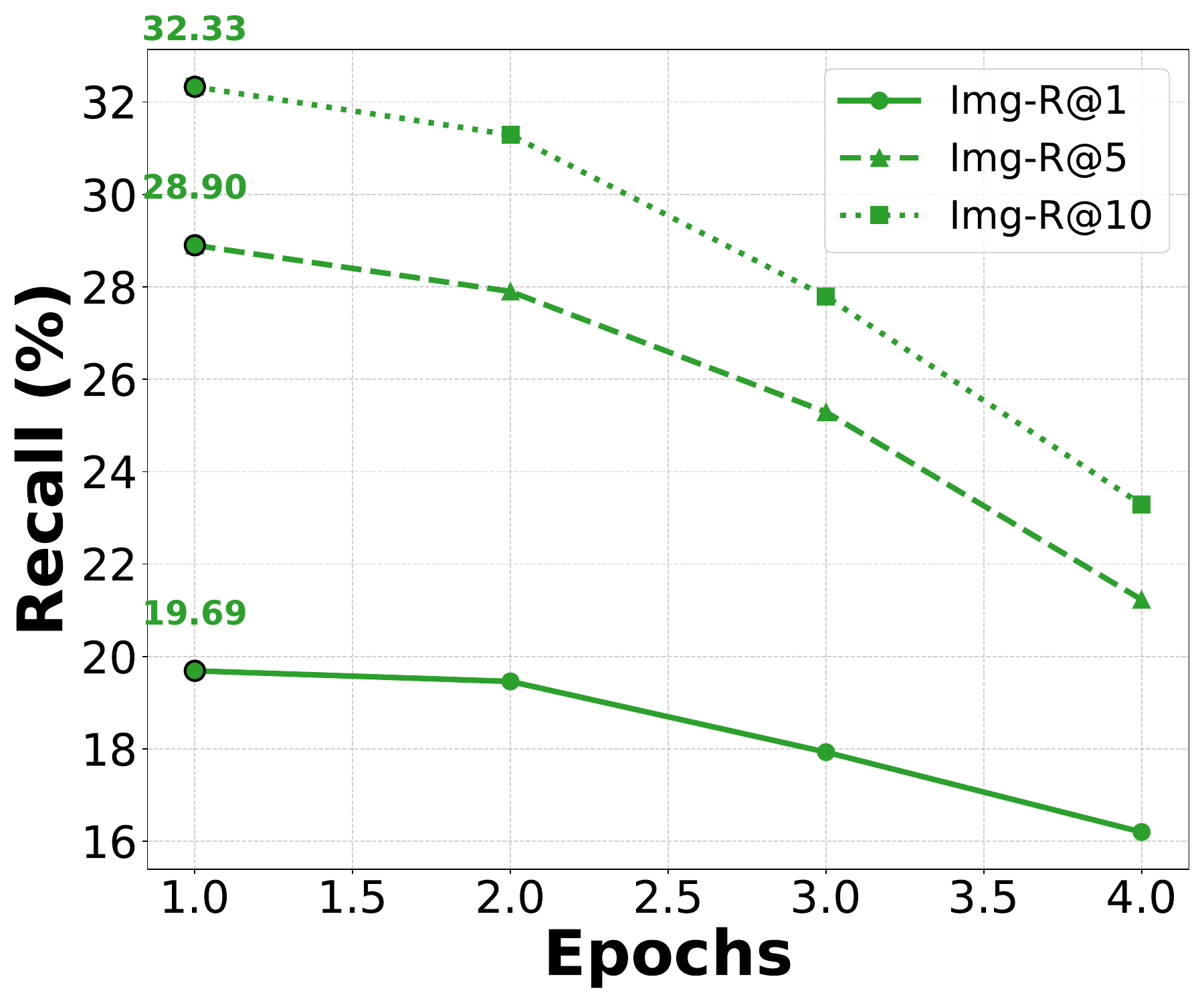}
        \caption*{(c) SciCap }
    \end{minipage}
    \begin{minipage}{0.24\textwidth}
        \centering
        \includegraphics[width=\linewidth]{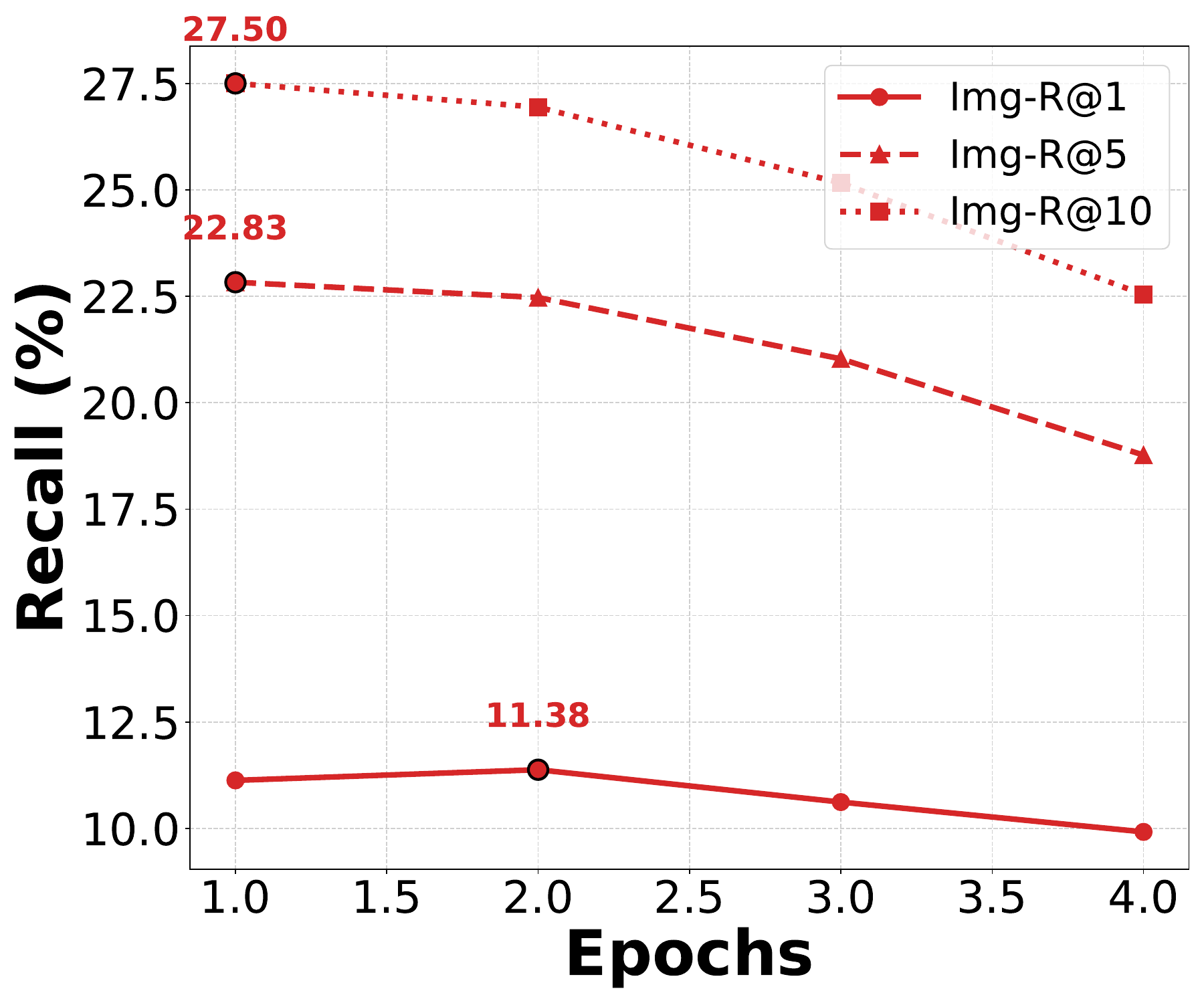}
        \caption*{(d) ArtCap }
    \end{minipage}
    \caption{Performance across 4 epochs of training. The results indicate that a single-step update yields optimal recall score overall.}
    \label{fig:performance_across_epochs}
\end{figure*}


\subsection{Top-k Selection}

The results in Table \ref{tab:topk_recall_comparison} show the impact of varying  the value of $k$ in the selection of the top candidates which form the few-shot training pool.
Increasing the top-$k$ value enhances recall performance, especially for higher recall metrics like Recall@10. For example, on the NASA dataset, Recall@10 improves from 67.95 at $k=8$ to 73.73 at $k=32$. 
On the one hand, this result is not surprising: 
with a higher $k$ the likelihood of including the ground truth image in the candidate pool is higher. On the other hand, without the improved re-ranking offered by EFSA, this would have no positive impact on the recall scores.

The Recall@1 metric exhibits a more nuanced pattern: while the scores initially improve as the top-$k$ value increases, the gains plateau or even decline beyond a certain top-$k$. For example, on COCO, Recall@1 reaches a peak at $k=16$ with a score of 30.14 but drops slightly as $k$ continues to increase. This suggests that while expanding the retrieval pool improves general recall, very large top-$k$ values may introduce additional noise,  compromising precision at the top rank.
Importantly, since performance gains tend to saturate at moderate $k$ values, one can default to a lower $k$ (e.g., 8) in compute-constrained settings without significant performance degradation. 

\subsection{Effect of Epoch}

All experiments so far were performed with a single epoch of training. In Figure \ref{fig:performance_across_epochs} we explore whether increasing the number of epochs has a positive impact on performance, with the finding that by and large that is not the case. For two datasets, COCO and ArtCap, we see slightly higher scores at Recall@1 and Recall@5 on the second epoch of training, but the general trend is for recall to drop with more extensive training. Interestingly, the drop is more pronounced at higher ranks. In NASA, for example, Recall@1 drops by less than 10 points across the four epochs of training,  while Recall@5 drops by over 25 points.
This indicates that prolonged training on a limited or domain-specific set of images can cause the model to memorize specific features rather than develop more generalized representations, robust across various types of images. Regardless, from a computational point of view, having to perform a single epoch of training to reap the benefits of EFSA, is optimal.

\subsection{Effect of Image Captioner}

In Table \ref{tab:llava_version} we measure EFSA's performance with captions generated with LLaVA-13B, LLaVA-7B \cite{liu2023llava} and TinyLLaVA \cite{zhou2024tinyllavaframeworksmallscalelarge}, and find that the choice of captioning model is not critical. Lighter models can be used to reduce computational overhead without a substantial change in results. To further assess the effectiveness of the synthetic captions, we compare their retrieval performance against ground-truth captions. Specifically, in the episodic fine-tuning we replace the synthetic captions with the ground-truth captions to obtain the results shown in Row 1 of Table~\ref{tab:llava_version}. The minimal gain in performance confirms that  synthetic captions effectively capture image semantics and serve as reliable substitutes for ground-truth captions.

\begin{table}[t]
\centering
\caption{Retrieval performance using different captioning models.}
\resizebox{\linewidth}{!}{%
\begin{tabular}{lccccccccc}
\toprule
\multirow{2}{*}{\textbf{Captions from}} & \multicolumn{3}{c}{\textbf{COCO}} & \multicolumn{3}{c}{\textbf{ArtCap}} \\
\cmidrule(lr){2-4} \cmidrule(lr){5-7}
 & \textbf{R@1} & \textbf{R@5} & \textbf{R@10} & \textbf{R@1} & \textbf{R@5} & \textbf{R@10} \\ 
\midrule
\rowcolor{blue!15} Ground-truth & 40.64 & 64.64 & 72.61  & 19.92 & 38.29 & 45.31 \\
LLaVA-13B   & \textbf{40.41} & \textbf{65.01} & \textbf{72.89} & 19.93 & \textbf{38.49} & \textbf{45.91} \\
 LLaVA-7B     & 40.33 & 64.96 & 72.69 & \textbf{19.95} & 38.40 & 45.71\\
TinyLLaVA-3.1B & 39.47 & 64.39 & 72.79 & 19.38 & 37.56 &45.18\\
\bottomrule
\end{tabular}
}
\label{tab:llava_version}
\end{table}






\section{Conclusion}
\label{sec:conclusion}

In this paper, we argue that the text-to-image retrieval performance of vision-language model should be evaluated in a multi-domain setting, characterized by a highly diverse pool of candidate images. Considering the limitations of zero-shot and finetuning methods in this context, we propose a novel Episodic Few-Shot Adaptation (EFSA) method, designed to enhance robustness against hard negatives in open-domain text-to-image retrieval tasks. By leveraging the top-$k$ candidate images along with synthetic captions generated for them, EFSA dynamically adapts to both domain- and sample-specific features, used to re-rank the top candidates and bring the ground-truth image to the very first rank. This approach consistently outperforms traditional fine-tuning and strong baselines across various benchmarks, demonstrating its effectiveness in mitigating domain-specific challenges and distributional shifts.



\clearpage

{
    \small
    \bibliographystyle{ieeenat_fullname}
    \bibliography{main}
}
\newpage

\clearpage
\setcounter{page}{1}
\maketitlesupplementary

In the following sections, we present additional results and a more extensive qualitative evaluation.

\section{Experiments with SigLIP }
\label{sec:siglib}

In Table \ref{tab:siglip_retrieval_comparison}, we provide multi-domain results with a more recent and performant vision-language model, SigLIP (ViT-SO400M-14) \cite{zhai2023sigmoid}. The strength of SigLIP over CLIP is evident in the Recall@k scores for the zero-shot baseline, all over 10 points higher for SigLIP compared to CLIP (see Table~\ref{tab:multi_domain}). 
Even with this stronger backbone, EFSA proves effective, yielding highest Recall@1 scores on 7 out of 8 datasets, the odd one out being yet again the Books dataset. That being said, the improvement is less pronounced here compared to the CLIP setting: the average Recall@1 increases from 34.48 to 36.15.
We hypothesize that the smaller performance gain is attributable to the stronger and more robust SigLIP backbone, which is inherently better at handling hard negatives.

\section{Effect of Caption Generation Prompts}
\label{supp:prompts}

\begin{figure}[h]
    \centering
    \includegraphics[height=2.3cm,width=\linewidth]{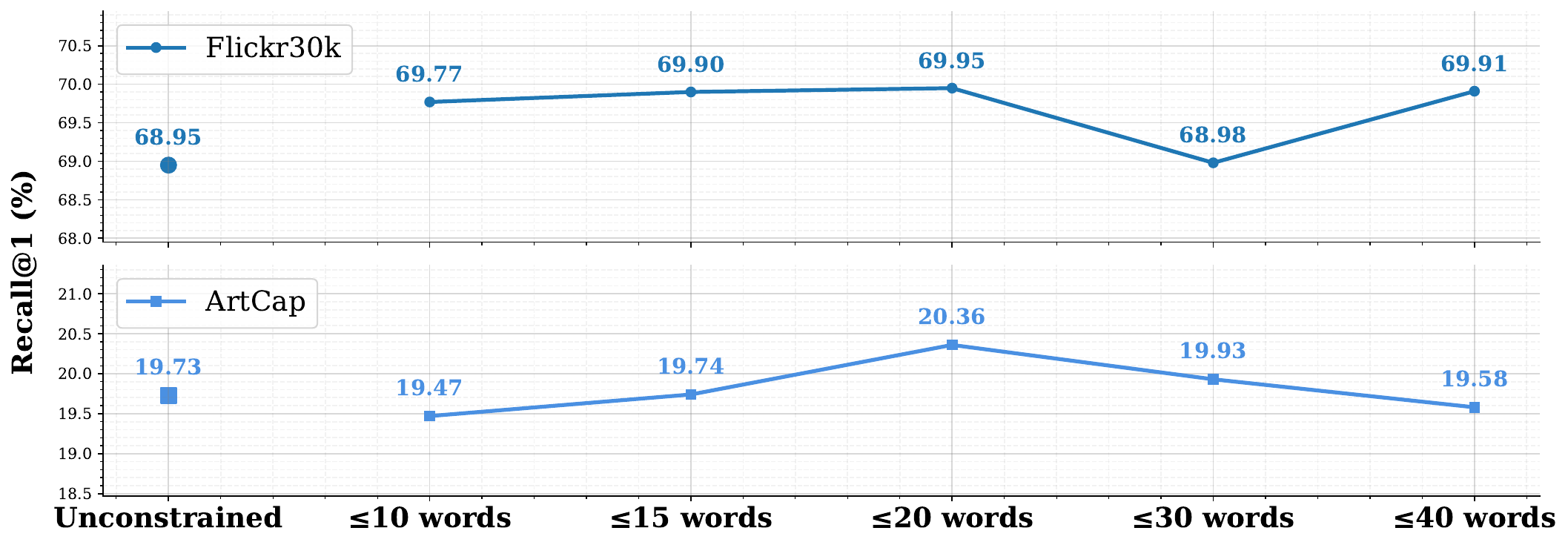}
    \caption{Effects of various caption generation prompts.}
    \label{fig:ece}
\end{figure}

Figure~\ref{fig:ece} shows how retrieval performance changes with different prompts for generating image captions. We tested prompts that varied in length constraints, from no word limit to a maximum of 10, 20, 30, or 40 words. Overall, the choice of prompt has less than 1 point impact across Flickr30k and ArtCap.
Performance improves when captions increase from 10 to 20 words but starts to decline as the word count goes beyond 20. 

\section{Qualitative Analysis}
\label{supp:qualitative}

\begin{figure*}
    \centering
    \begin{subfigure}[]{0.94\linewidth}
        \includegraphics[width=\textwidth]{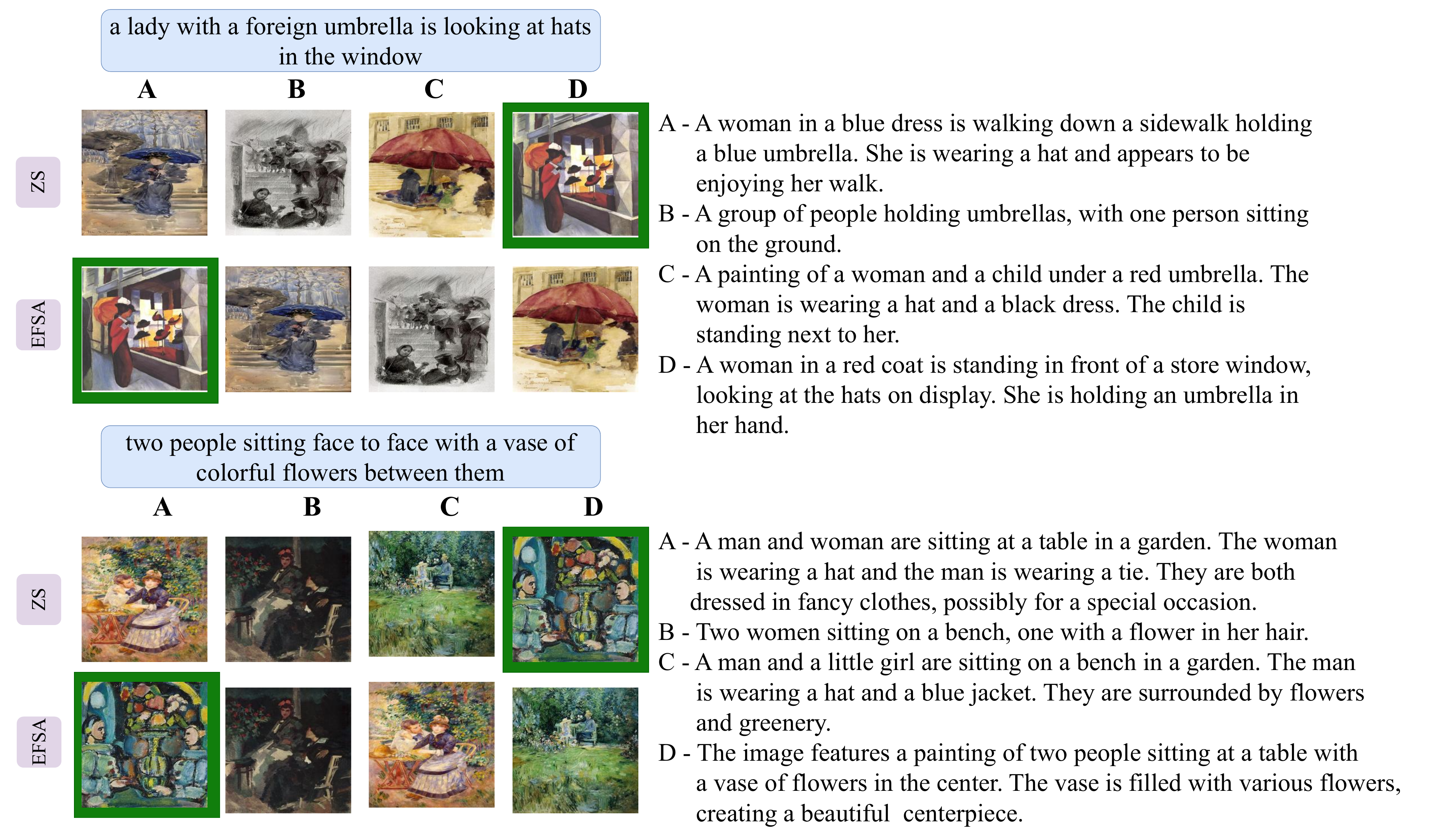}
        \caption{Qualitative examples from ArtCap.}
    \end{subfigure}
    \begin{subfigure}[b]{0.94\linewidth}
        \includegraphics[width=\textwidth]{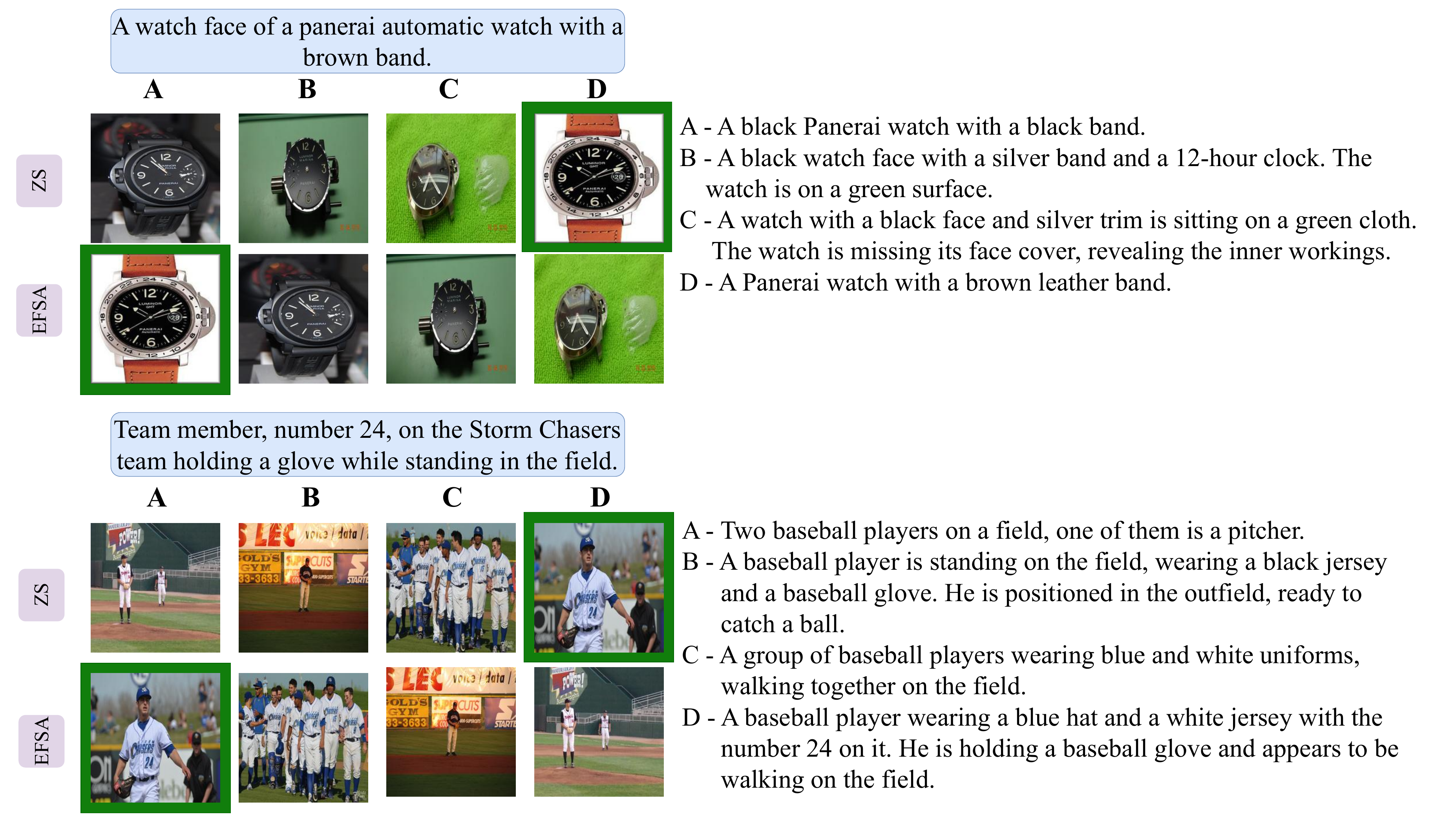}
        \caption{Qualitative examples from TextCap.}
    \end{subfigure}
    \caption{Qualitative comparison between EFSA and zero-shot CLIP on the ArtCap (top teo examples) and TextCap (bottom two examples) datasets in the single-domain setting. Green-framed images indicate the ground-truth for each text query, displayed on top. EFSA effectively re-ranks the ground-truth images to the top rank, outperforming zero-shot CLIP. On the right, the synthetic caption for each image is provided, as used for episodic few-shot adaptation.}
    \label{fig:artcap_textcap_sup}
\end{figure*}

Figure~\ref{fig:artcap_textcap_sup} presents a qualitative comparison between zero-shot CLIP and EFSA in terms of the top-4 retrieved images on the ArtCap and TextCap datasets, with the synthetic captions for the images also included. We observe that the synthetic captions exhibit considerable semantic overlap with the query text. Notably, LLaVA accurately interprets text present within the images and incorporates it into the captions.  Yet, as discussed in \S\ref{sec:results}, a simple text-to-text retrieval approach does not prove effective here. EFSA instead enables the backbone model to learn from the image-caption pairs, leveraging not only information from the ground-truth image but also from the hard negatives surrounding it. Using this information to build more accurate representations for the images in the retrieval pool, the EFSA-modified CLIP can correclty re-rank the ground-truth image to the top position.

\section{Storage Overhead Analysis}
\label{supp:storage}

We compare the storage requirements introduced by EFSA—due to storing additional top-$k$ captions—with the zero-shot baseline that stores only image embeddings.

\paragraph{Baseline Storage:}  
We estimate the storage requirement of the zero-shot T2I baseline as:  
\begin{align}
\texttt{Total Storage} &= \langle \textit{size of retrieval pool} \rangle \nonumber \\
&\quad \times \langle \textit{bytes per cached image rep} \rangle.
\end{align}
Each cached image embedding is a 768-dimensional \texttt{float32} vector (as used in CLIP-base), requiring:
\begin{equation}
768 \times 4 = 3072 \ \text{bytes per image}.
\end{equation}

\paragraph{EFSA Caption Overhead:}  
EFSA adds one generated caption per image, stored as a sequence of CLIP vocabulary token IDs, each encoded as a \texttt{uint16}. Assuming an average of 30 tokens per caption, the additional storage required is:
\begin{equation}
30 \ \text{tokens} \times 2 \ \text{bytes/token} = 60 \ \text{bytes per image}.
\end{equation}

\paragraph{Relative Overhead:}  
The relative increase in storage per image is given by:
\begin{equation}
\frac{60}{3072} \approx 0.0195 \ (\approx 2\%).
\end{equation}

Thus, EFSA introduces a minimal storage overhead of approximately 2\% compared to the zero-shot baseline, while providing measurable gains in both open- and closed-domain retrieval performance.

\begin{table*}
\fontsize{10pt}{5pt}\selectfont
\centering
\caption{Text-to-image retrieval performance in a multi-domain setting \textbf{with a SigLIP backbone}. Results are reported for Zero-Shot (Z.S), Fine-Tuning (F.T), Text-to-Text (T2T), and Episodic Few-Shot Adaptation (EFSA). The results demonstrate that EFSA consistently surpasses other methodologies, particularly on Recall@1 in this complex retrieval setup. }
\renewcommand{\arraystretch}{1.4}
\resizebox{\linewidth}{!}{%
\begin{tabular}{lcccccccccccccccccccccccc}
\toprule
& \multicolumn{12}{c}{\textbf{Multi-domain}}\\
\midrule
\multirow{2}{*}{} & \multicolumn{3}{>{\columncolor{gray!20}}c}{\textbf{COCO}} & \multicolumn{3}{>{\columncolor{gray!20}}c}{\textbf{Flickr30k}} & \multicolumn{3}{>{\columncolor{gray!20}}c}{\textbf{Books}} & \multicolumn{3}{>{\columncolor{gray!20}}c}{\textbf{NASA}} \\ 
\cmidrule(lr){2-4}
\cmidrule(lr){5-7}
\cmidrule(lr){8-10}
\cmidrule(lr){11-13}
& \cellcolor{gray!10}R@1 & R@5 & R@10 & \cellcolor{gray!10}R@1 & R@5 & R@10 & \cellcolor{gray!10}R@1 & R@5 & R@10 & \cellcolor{gray!10}R@1 & R@5 & R@10  \\

\cmidrule(lr){1-13}

Z.S & \cellcolor{gray!10}39.1 & 61.71 & 70.31 & \cellcolor{gray!10}49.23 & 72.15 & 79.29 & \cellcolor{gray!10}\textbf{32.27} & \textbf{49.98} & \textbf{54.9} & \cellcolor{gray!10}14.9 & 27.22 & \textbf{35.66} \\
  F.T & \cellcolor{gray!10}30.23 & 53.32 & 63.21 & \cellcolor{gray!10}34.74 & 58.89 & 67.72 & \cellcolor{gray!10}7.72 & 16.3 & 20.89 & \cellcolor{gray!10}3.61 & 10.12 & 13.73 \\
T2T & \cellcolor{gray!10}18.27 & 32.50 & 39.27 & \cellcolor{gray!10}20.44 & 33.00 & 39.46 & \cellcolor{gray!10}0.98 & 2.05 & 2.75 & \cellcolor{gray!10}2.16 & 4.81 & 6.26 \\
\rowcolor{blue!15} EFSA & \textbf{42.61} & \textbf{64.69} & \textbf{72.27} & \textbf{52.49} & \textbf{75.08} & \textbf{80.74} & 31.55 & 48.44 & 53.84 & \textbf{15.18} & \textbf{27.46} & 34.69 \\

\midrule

\multirow{2}{*}{} & \multicolumn{3}{>{\columncolor{gray!20}}c}{\textbf{VizWiz}} & \multicolumn{3}{>{\columncolor{gray!20}}c}{\textbf{TextCap}} & \multicolumn{3}{>{\columncolor{gray!20}}c}{\textbf{ArtCap}} & \multicolumn{3}{>{\columncolor{gray!20}}c}{\textbf{SciCap}} & \multicolumn{3}{>{\columncolor{gray!20}}c}{\textbf{Average}}   \\ 
\cmidrule(lr){2-4}
\cmidrule(lr){5-7}
\cmidrule(lr){8-10}
\cmidrule(lr){11-13}
\cmidrule(lr){14-16}

& \cellcolor{gray!10}R@1 & R@5 & R@10 & \cellcolor{gray!10}R@1 & R@5 & R@10 & \cellcolor{gray!10}R@1 & R@5 & R@10 & \cellcolor{gray!10}R@1 & R@5 & R@10 & \cellcolor{gray!10}R@1 & R@5 & R@10 \\

\midrule

Z.S & \cellcolor{gray!10}31.99 & 49.24 & 55.55 & \cellcolor{gray!10}58.75 & 73.00 & 77.93 & \cellcolor{gray!10}13.21 & 28.67 & 36.89 & \cellcolor{gray!10}36.46 & \textbf{50.49} & \textbf{56.53} & \cellcolor{gray!10}34.48 & 51.55 & 58.38\\
F.T & \cellcolor{gray!10}22.98 & 39.43 & 46.54 & \cellcolor{gray!10}43.74 & 60.08 & 66.34 & \cellcolor{gray!10}10.65 & 24.03 & 31.87 & \cellcolor{gray!10}5.96 & 12.33 & 15.56 & \cellcolor{gray!10}19.95 & 34.31 & 40.73\\
T2T & \cellcolor{gray!10}14.07 & 24.28 & 28.90 & \cellcolor{gray!10}22.68 & 33.32 & 38.40 & \cellcolor{gray!10}5.33 & 12.27 & 16.41 & \cellcolor{gray!10}8.63 & 13.79 & 16.69 & \cellcolor{gray!10}11.57 & 19.50 & 23.51 \\

\rowcolor{blue!15} EFSA & \textbf{33.66} & \textbf{50.85} & \textbf{56.28} & \textbf{60.95} & \textbf{74.52} & \textbf{78.94} & \textbf{15.45} & \textbf{31.43} & \textbf{38.52} & \textbf{37.36} & 50.33 & 55.4 & \textbf{36.15} & \textbf{52.85} & \textbf{58.83} \\

\bottomrule
\end{tabular}%
}
\label{tab:siglip_retrieval_comparison}
\end{table*}

\section{Effect of Loss Function}
\label{supp:loss}

\begin{table}[]
\centering
\caption{Effect of various loss Functions on text-to-image Retrieval performance. A weighted combination of contrastive and hinge loss enhances retrieval performance.}
\resizebox{\linewidth}{!}{%
\begin{tabular}{lccccccccc}
\toprule
\multirow{2}{*}{\textbf{Loss Function}} & \multicolumn{3}{c}{\textbf{COCO}} & \multicolumn{3}{c}{\textbf{ArtCap}} \\
\cmidrule(lr){2-4} \cmidrule(lr){5-7}
 & \textbf{R@1} & \textbf{R@5} & \textbf{R@10} & \textbf{R@1} & \textbf{R@5} & \textbf{R@10} \\ 
\midrule

Hinge   & \textbf{30.15} & 50.78 & 57.79 & 10.91 & 22.73 & 27.47 \\
 Contrastive     & 28.23 & 49.20 & 56.47 & 10.33 & 21.66 & 26.61 \\
\rowcolor{blue!15} Combined & 30.14 & \textbf{50.96} & \textbf{57.82} & \textbf{11.13} & \textbf{22.83} & \textbf{27.50} \\
\bottomrule
\end{tabular}
}
\label{tab:loss_comparison}
\end{table}

Table \ref{tab:loss_comparison} shows the impact of different training objectives on retrieval performance across the COCO and ArtCap datasets. Hinge loss proves more effective than the contrastive loss here, likely due to the nature of the data, consisting of highly-similar images with highly similar captions. Through the margin in the hinge loss, these data points are being actively pushed away from each other, forcing the model to pay attention to subtle differences. 
Although the contrastive loss is less performant than the hinge loss on its own, the combination of the two yields the best performance on average, the difference being more pronounced in the more complex ArtCap domain.

\section{Impact of False Negatives}

\noindent
In an open-domain setting, a given query $x$ may have multiple valid image matches $Y$, even though our evaluation setup only designates a single ground-truth. For instance, queries from COCO may have equally valid counterparts in closely related datasets such as Flickr, which are then characterized as \textit{false negatives}. This ambiguity makes evaluation nuanced. Nevertheless, EFSA consistently improves Recall@1 with respect to the designated ground-truth targets, suggesting that the false negative issue, while possible, is not ubiquitous. In cases where it does interfere with the interpretation of Recall@1, we can additionally rely on Recall@5 and Recall@10, which reflect whether the ground-truth is still being successfully promoted to the top ranks despite the presence of potential false negatives.

\label{sec:rationale}

\end{document}